\documentclass[11pt]{article}

\usepackage[preprint]{acl}

\usepackage{times}
\usepackage{latexsym}

\usepackage[T1]{fontenc}

\usepackage[utf8]{inputenc}

\usepackage{microtype}

\usepackage{inconsolata}

\usepackage{graphicx}

\usepackage{amsmath}   
\usepackage{amsfonts}  
\usepackage{amssymb}
\usepackage{booktabs}
\usepackage{caption}
\usepackage{tabularx,booktabs}
\usepackage{stfloats}
\usepackage{float}
\usepackage{multirow}
\usepackage{graphicx}
\graphicspath{{figs/}}
\usepackage{listings}
\usepackage{subcaption}

\usepackage[utf8]{inputenc}   
\usepackage[T1]{fontenc}
\usepackage{xcolor}
\usepackage{listings}
\usepackage{hyperref} 
\usepackage{xurl}

\usepackage{listings,xcolor}
\usepackage{authblk}

\title{The Gaining Paths to Investment Success: Information-Driven LLM Graph Reasoning for Venture Capital Prediction}

\author{
  Haoyu Pei\textsuperscript{1},
  Zhongyang Liu\textsuperscript{1},
  Xiangyi Xiao\textsuperscript{1},
  Xiaocong Du\textsuperscript{1} \\
  {\normalfont\normalsize
  Suting Hong\textsuperscript{2$\dagger$},
  Kunpeng Zhang\textsuperscript{3$\dagger$},
  Haipeng Zhang\textsuperscript{1$\dagger$}} \\
  \textsuperscript{1}ShanghaiTech University \quad
  \textsuperscript{2}Xi'an Jiaotong-Liverpool University \quad
  \textsuperscript{3}University of Maryland \\
  \texttt{\{peihy2024, liuzy12024, xiaoxy, duxc2023, zhanghp\}@shanghaitech.edu.cn} \\
  \texttt{sutinghong@gmail.com} \quad
  \texttt{kpzhang@umd.edu}
}

\begin{document}

\maketitle
\begin{abstract}

Most venture capital (VC) investments fail, while a few deliver outsized returns. Accurately predicting startup success requires synthesizing complex relational evidence—company disclosures, investor track records, and investment network structures—through \emph{explicit reasoning} to form coherent, interpretable investment theses. Traditional machine learning and graph neural networks both lack this reasoning capability. LLMs offer strong reasoning but face a modality mismatch with graphs. Recent graph-LLM methods target \emph{in-graph} tasks where answers lie within the graph, whereas VC prediction is \emph{off-graph}: the target exists outside the network. The core challenge becomes selecting graph paths that maximize predictor performance on an external objective while enabling step-by-step reasoning. We present MIRAGE-VC, addressing two obstacles: \emph{path explosion} (thousands of candidate paths overwhelm LLM context) and \emph{heterogeneous evidence fusion} (different startups need different analytical emphasis). Our information-gain-driven path retriever iteratively selects high-value neighbors, distilling investment networks into compact chains for explicit reasoning. A multi-agent architecture integrates three evidence streams via learnable gating based on company attributes. Under strict anti-leakage controls, MIRAGE-VC achieves +5.0\% F1 and +16.6\% Precision@5, and sheds light on other off-graph prediction tasks such as recommendation and risk assessment. Our code is available.\footnote{\url{https://anonymous.4open.science/r/MIRAGE-VC-323F}}

\end{abstract}

\section{Introduction}

Venture capital (VC) investment is characterized by extreme asymmetry: from 1985 to 2009, roughly 60\% of VC-backed firms lost money, while only 10\% returned over five times the initial investment~\cite{kerr2014entrepreneurship}. This high-risk, high-reward profile makes accurate startup success prediction crucial for portfolio optimization and capital allocation.

\begin{figure}[t] 
  \centering
  \includegraphics[width=1.0\linewidth]{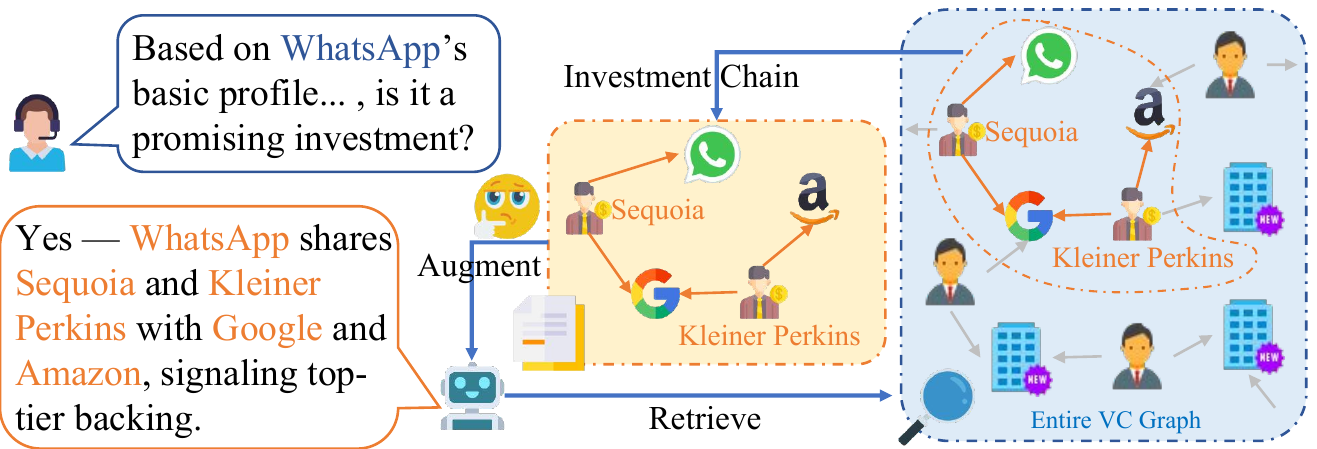} 
  \caption{Impact of the selected chain on prediction.}
  \label{fig:intro}
\end{figure}

The challenge lies in synthesizing complex relational evidence across multiple sources~\cite{gompers2020venture}. A typical investment decision requires integrating (i) the target company's peer-relative market positioning~\cite{kim1999valuing}, (ii) the track record and reputation of its lead investors~\cite{hsu2004entrepreneurs}, and (iii) structural signals embedded in the investment network—such as co-investment patterns, investor coalitions, and proximity to successful exits~\cite{hochberg2007whom}. For instance, when evaluating a messaging startup like WhatsApp, the investment chain ``WhatsApp $\leftarrow$ Sequoia $\rightarrow$ Google $\leftarrow$ Kleiner Perkins $\rightarrow$ Amazon'' (Figure~\ref{fig:intro}) indicates: top-tier screening by Sequoia increases follow-on financing probability, proximity to tech giants enhances scalability potential, and coalition strength among premier VCs facilitates strategic M\&A opportunities. Integrating such heterogeneous evidence requires not only pattern recognition but also \emph{explicit reasoning}—connecting the dots across comparables, investor profiles, and graph structures to form coherent, interpretable investment theses~\cite{kaplan2004characteristics}. 

Existing predictive approaches fall short on these requirements. Traditional machine learning methods~\cite{arroyo2019assessment, bento2017predicting} rely on isolated firm features and ignore relational context. Graph neural networks (GNNs)~\cite{lyu2025help, zhang2021scalable} capture high-order investor–company dependencies but operate as a black box without exposing underlying reasoning. Consequently, they lack interpretability—providing no human-readable explanation for why a startup is promising—and cannot incorporate external knowledge like recent funding trends 
outside the training graph.

Large language models (LLMs) offer a natural remedy through their strong reasoning and broad world knowledge~\cite{liu2023fingpt, ko2024can}. However, LLMs face a fundamental \emph{modality mismatch} when processing graph-structured data~\cite{wang2023can}: their architectures are optimized for sequential text, not relational topologies. While recent work~\cite{sun2023think, luo2023reasoning} integrates LLMs with knowledge graphs (KGs) for \emph{in-graph question answering}—where the answer is a node within the graph (e.g., ``Which company did Sequoia invest in after Google?'')—VC prediction is an \emph{off-graph} task: the prediction target (startup success) exists outside the network, and the graph serves solely as retrievable evidence. This distinction applies to other domains such as recommendation systems (predicting user–item affinity from interaction graphs) and credit risk assessment (predicting default from transaction networks).
 The core challenge thus becomes: \textbf{\emph{how to select graph-derived paths that maximize a predictor's performance on an external objective}}, rather than finding paths that terminate at an in-graph answer.

Even if we restrict our focus to path-based retrieval, the primary obstacle is \textbf{exponential path explosion}: shallow 1-hop neighborhoods around a target company often lack sufficient signal~\cite{yu2021recognizing}, yet extending to 3 or 4 hops generates thousands of candidate paths, many of which are redundant or weakly informative~\cite{zhang2025diagnosing}. Indiscriminately feeding all paths to an LLM overwhelms its context window and dilutes attention. Additionally, VC prediction requires integrating semantically heterogeneous sources—company disclosures, investor profiles, and graph paths—whose relative importance varies by startup type, demanding adaptive fusion mechanisms to avoid over-reliance on secondary signals.

To address these challenges, we propose MIRAGE-VC, a multi-perspective retrieval-augmented generation (RAG) framework for VC prediction. To tackle path explosion, we introduce an \textbf{\emph{information-gain-driven path retriever}} that iteratively expands from the target company, at each hop selecting the neighbor whose inclusion maximally improves an LLM predictor's accuracy. This selector is trained offline using task-specific information gain signals computed by a frozen LLM, then deployed at inference to extract a compact set of high-value investment chains—enabling explicit, chain-of-thought reasoning without overwhelming the LLM's context window. To integrate heterogeneous evidence, we employ a \textbf{\emph{learnable gating network}} that dynamically weights verdicts from three information streams—company disclosures, lead investor profiles, and graph-based paths—each analyzed by a dedicated LLM agent. A manager agent then synthesizes these weighted signals into a calibrated prediction with an interpretable rationale.

Our contributions are as follows:
\begin{itemize}
    \item By distilling complex VC networks into high-gain paths that synthesize evidence across multiple investors and deals, we allow LLMs to reason over relational patterns, exposing actionable investment signals valuable to VC practitioners.
    \item Our gating mechanism dynamically weights heterogeneous evidence streams based on startup characteristics, mirroring how different venture types require different analytical emphasis.
    \item Under strict anti-leakage controls, we achieve state-of-the-art performance on real-world VC data (+5.0\% F1, +16.6\% Precision@5). Our paradigm of selecting graph paths by marginal utility generalizes to other off-graph prediction tasks, such as recommendation and risk assessment.
\end{itemize}

\section{Related Work}
\subsection{Graph-based VC Prediction}
Traditional machine learning predictors rely on independent firm-level features and ignore relational context~\cite{arroyo2019assessment, bento2017predicting}, whereas GNNs model investor–company graphs to capture high-order relational signals.
SHGMNN~\cite{zhang2021scalable} combines predefined meta-paths, lightweight GNNs and Markov random field inference to integrate heterogeneous topologies and propagate labels for large-scale early-stage startup identification. GST~\cite{lyu2025help} applies unsupervised graph self-attention to update a dynamic startup-investor bipartite graph, improving node embeddings via link prediction and node classification to capture rich investor–company relations. These studies demonstrate that the structural properties of VC investment networks can significantly improve predictive accuracy. However, they remain limited by narrow knowledge scopes, weak reasoning capabilities, and a lack of interpretability.

\subsection{Graph-augmented LLMs}
Recent systems couple LLMs with knowledge graphs for \emph{in-graph} QA—answers are entities in graph, reached or verified via triple retrieval and path reasoning~\cite{sun2023think,luo2023reasoning}.
While effective at fact verification, they optimize in-graph objectives (entity/relation correctness) rather than selecting multi-hop evidence by its marginal utility to an external predictor.
In our off-graph VC setting, the graph serves as evidence for startup-success prediction; what is needed is utility-aware selection of a few high-value investment chains as explicit evidence for prediction.

Proposed to remedy text-only RAG’s inability to model structure and multi-hop dependencies, GNN-RAG frameworks retrieve relevant nodes via embedding similarity and inject local structural cues before handing context to an LLM~\cite{mavromatis2024gnn}.
Yet they typically surface no explicit reasoning paths, limiting LLMs’ strength in stepwise, interpretable chain-of-thought reasoning~\cite{wei2022chain} and constraining multi-hop inference over heterogeneous investment networks.

\begin{figure*}[t]  
  \centering
  \includegraphics[width=\textwidth]{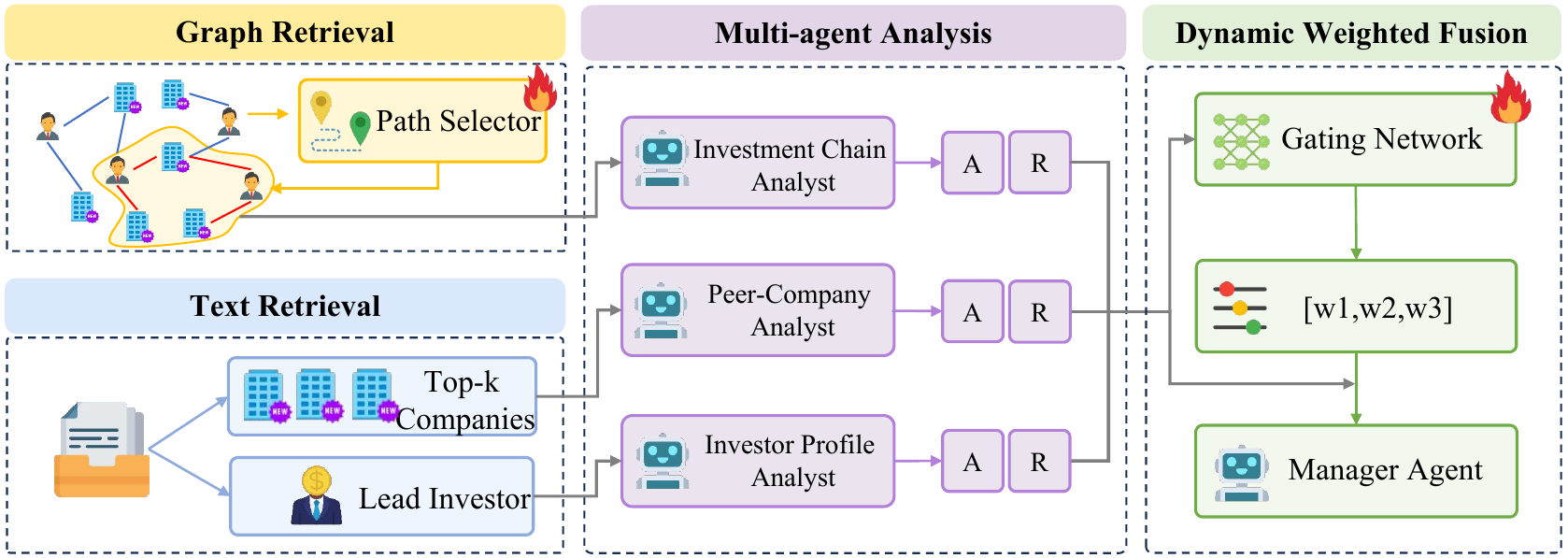}
  \caption{Overall Framework of MIRAGE-VC. It contains four key components: \emph{Graph Retrieval} for path selecting, \emph{Text Retrieval} for companies and investor, \emph{Multi-agent Analysis} for multi-perspective information, \emph{Dynamic Weighted Fusion} for adaptive information fusion.}
  \label{fig:method-overview}
\end{figure*}
\section{Preliminary}
\subsection{Problem definition}
This study aims to predict the success of early-stage start-ups, defined as companies that have completed their first formal financing round (seed or angel) but have not yet raised Series A funding~\cite{zhang2021scalable}. While success is often measured by the attainment of Series A financing, prior studies use varying observation windows, which can introduce temporal bias. To mitigate this, we adopt a consistent one-year observation window following the seed round. This approach aligns with stage-based evaluation practices and helps control for external environmental factors~\cite{boocock1997evaluation}. The core task is to predict whether an early stage startup will secure subsequent financing within one year of its initial funding.

\subsection{Data overview}
We use the PitchBook\footnote{PitchBook is a financial data platform providing comprehensive information on private and public capital markets, including venture capital, private equity, and M\&A transactions.} Global VC dataset, which spans investment activities from 2005 to November 2023. The dataset includes detailed investment records specifying the invested company, investor identity, funding amount, and financing stage. It also contains demographic information on both entrepreneurs and investors, including background, location, education, and professional biographies. Additionally, startup-level attributes are provided, such as team composition, industry classification, keyword tags, and geographic location. In total, the dataset encompasses 263,729 startups and 1,014,157 individuals. See Appendix~\ref{app:data} for more details.

\subsection{VC investment network}
We model the VC ecosystem as a time‐stamped heterogeneous information network \(G=(\mathcal{V},\mathcal{E})\), where \(\mathcal{V}=\mathcal{V}_{\mathrm{cmp}}\cup\mathcal{V}_{\mathrm{inv}}\) contains company and investor nodes. Each directed edge \(e=(v_{\mathrm{inv}},v_{\mathrm{cmp}},t)\) represents an investment event from investor to company at time \(t\), annotated with attributes such as the financing round and investment amount. For each company \(c^*\) that completes an angel or seed round at time \(t\), we assign a binary label \(y^*=1\) if it secures Series A funding within the following 12 months, and \(y^*=0\) otherwise.

\section{Methodology}
\subsection{Overview of our method}
As shown in Figure~\ref{fig:method-overview}, our proposed framework follows four sequential stages: (1) Graph retrieval: To supply the investment chain agent with structured evidence, a learnable graph retriever extracts a high-value company–investor path from the investment graph. (2) Text retrieval: To provide the company and investor agents with textual context, semantic matching over public filings yields two textual views: (i) a similar company context and (ii) a lead-investor profile composed of  demographics, career history, and labeled deal records. (3) Multi-agent analysis: The three prompts are processed by frozen LLM agents, each independently returning a binary decision and supporting rationale. (4) Perspective fusion: A lightweight gating network embeds and weighs the agent outputs. These are passed to a frozen manager agent, which produces a calibrated success probability and interpretable final decision.

\subsection{Graph Retrieval}
\subsubsection{From classic IG to graph paths}
\begin{figure}[t]
  \centering
  \includegraphics[width=\linewidth]{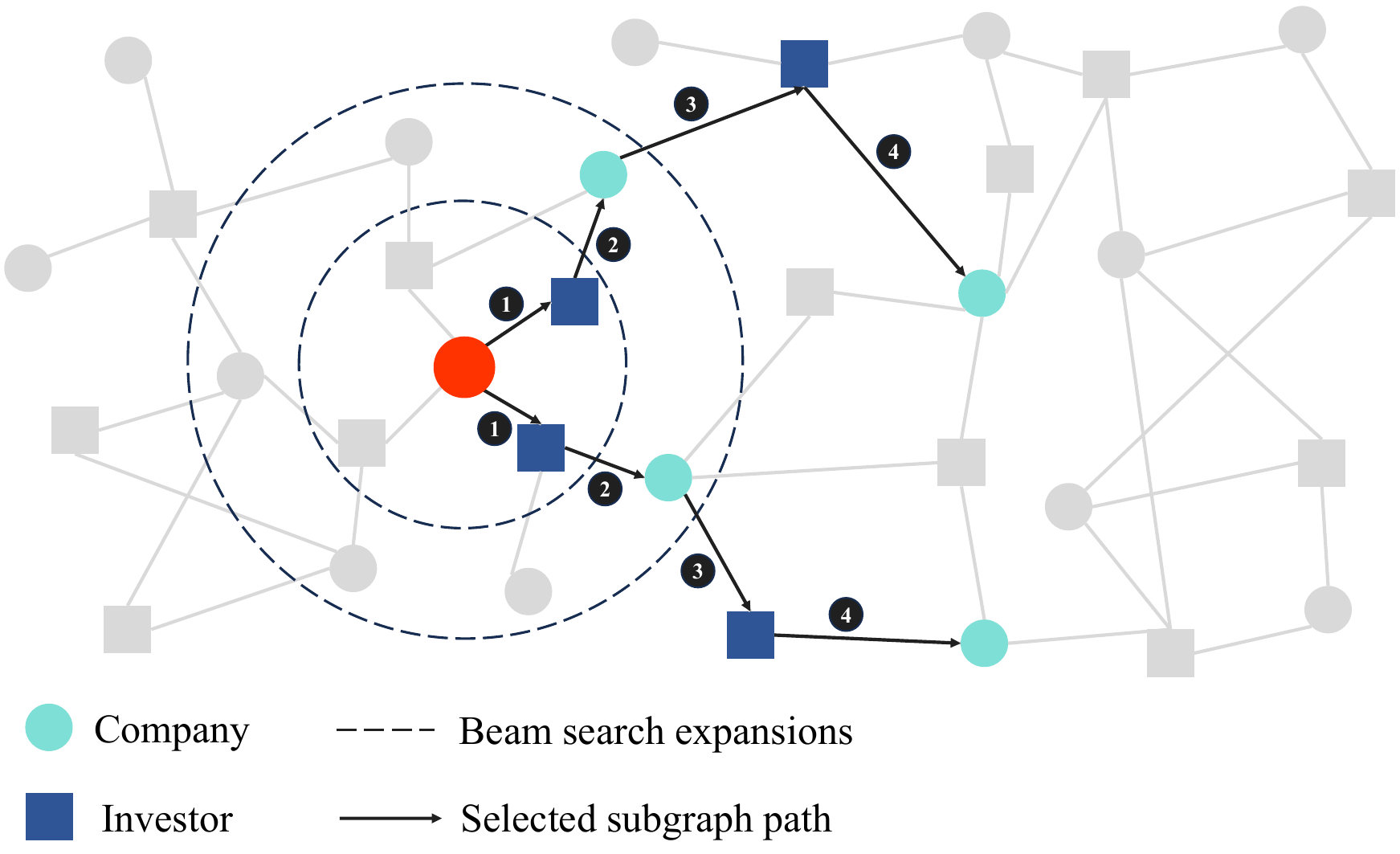}
  \caption{Illustration of how the path selector retrieves the best path from graph}
  \label{fig:acl_beamsearch}
\end{figure}

As shown in Figure~\ref{fig:acl_beamsearch}, path selection is framed as a sequential node selection problem. Starting from the target node \(c^{*}\), at each hop, we choose the neighbor whose inclusion maximally improves the model's prediction accuracy. Although this heuristic does not guarantee a globally optimal path, it provides an efficient approximation—analogous to decision-tree splits via information gain—well suited to our multi-hop retrieval setting~\cite{quinlan1986induction}:
\begin{equation}
\text{IG}(A) = H(Y) - H\bigl(Y \mid A\bigr)
\label{eq:ig}
\end{equation}
We extend this principle to graphs by treating each candidate node \(v\) as an ``attribute'' \(A\) and estimating label uncertainty using the cross-entropy of a frozen LLM predictor. The rest of this section describes how these LLM-based IG signals are annotated offline, and how a lightweight selector model is trained to approximate them during inference.

\subsubsection{LLM-generated gain labels}
To obtain oracle supervision for path selector, we use a frozen LLM to quantify task‐specific information gain per candidate expansion. For each target company \(c^{*}\), we build a breadth-first expansion tree of depth at most three, retaining up to three previously unseen neighbors per node. At hop \(h\in\{0,1,2\}\) let  \(S^{(h)}=\langle c^{*},\dots ,u\rangle\) denote a current path and \(\{v_{1},v_{2},v_{3}\}\subseteq N(u)\!\setminus\!S^{(h)}\) the
corresponding candidate set.

\paragraph{Prompt construction}
To measure the incremental value of each candidate node \(v_{i}\), we generate two prompts per expansion: (i) a \textit{baseline} prompt \(P_{\mathrm{base}}\) that verbalizes nodes in \(S^{(h)}\), and (ii) a \textit{candidate} prompt \(P_{v_{i}}\) that verbalizes the
extended path \(S^{(h)}_{v_{i}}
      =\langle c^{*},\dots ,u,v_{i}\rangle\).
A frozen \textsc{Llama-3.1-8B} classifier returns the success probabilities \(p_{\mathrm{base}}\) and \(p_{v_{i}}\). The procedure for converting causal-LM logits into binary probabilities \(p\) is detailed in Appendix.

\paragraph{Task-specific information gain}
Given the gold label
\(y\in\{0,1\}\) (1 = \textit{Success}, 0 = \textit{Failure}),
we define the marginal gain from including \(v_{i}\):
\begin{multline}
\Delta_{v_i}
= \underbrace{\mathrm{CE}\!\left(y,\,p_{\mathrm{base}}\right)
             - \mathrm{CE}\!\left(y,\,p_{v_i}\right)}_{\text{cross-entropy reduction}} \\
\quad
+ \lambda_{\mathrm{conf}}
  \bigl(\lvert p_{v_i}-0.5\rvert
       - \lvert p_{\mathrm{base}}-0.5\rvert \bigr)
\label{eq:delta_gain}
\end{multline}
where the first term rewards the reduction in the prediction error
(irrespective of \(y=0\) or \(1\)); the second encourages confidence once the correctness is taken into account. \(\mathrm{CE}\) denotes binary cross-entropy and \(\lambda_{\mathrm{conf}}\in[0,1]\) balances correctness against confidence shift.

\paragraph{Training tuples}
Each training instance consists of:
\(
\bigl(
   S^{(h)},\;
   S^{(h)}_{v_{i}},\;
   \Delta_{v_{i}}
 \bigr)
\)
The selector later receives the baseline path \(S^{(h)}\), the extended path \(S^{(h)}_{v_{i}}\), and the scalar gain \(\Delta_{v_{i}}\) it should learn to predict. Because gains are computed for both successful and failed companies, the selector is explicitly trained to prefer extensions that push the LLMs towards the correct class with higher confidence.

\subsubsection{Selector training objective}
Each hop $h$ of a target company contributes one \emph{ranking group}
\(G^{(h)}=\{v_{1},v_{2},v_{3}\}\) with associated gains
\(\Delta_{v_{1}},\Delta_{v_{2}},\Delta_{v_{3}}\)
annotated as in Eq.~\eqref{eq:delta_gain}.
For each candidate \(v\in G^{(h)}\), we compute a difference feature:
\begin{equation}
x_{v}
= \bigl[e_{\mathrm{base}}
      \;\Vert\;
      e_{v}
      \;\Vert\;
      (e_{v}-e_{\mathrm{base}})\bigr]
\;\in\;\mathbb{R}^{2304}
\label{eq:xv_representation}
\end{equation}
where \(e_{\mathrm{base}}\) and \(e_{v}\) are $768$-dimensional sentence embeddings extracted once by a frozen encoder.
A lightweight two-layer MLP \(s_{\theta}:\mathbb{R}^{2304}\!\to\!\mathbb{R}\)
assigns a score to each expansion.
\paragraph{Listwise objective}
To match the full gain pattern within each group we optimize a listwise objective. We first apply a within-group shift \(r_i=\Delta_{v_i}-\min_j \Delta_{v_j}\), which preserves ordering while ensuring non-negativity (\(r_i\!\ge\!0\) and \(\arg\max_i r_i=\arg\max_i \Delta_{v_i}\)). We then form temperature-smoothed targets
\begin{equation}
q_i=\frac{\exp(r_i/\tau)}{\sum_{j=1}^{k}\exp(r_j/\tau)}
\label{eq:list-soft-q}
\end{equation}
\begin{equation}
p_i=\frac{\exp(s_{\theta}(x_{v_i})/\tau)}{\sum_{j=1}^{k}\exp(s_{\theta}(x_{v_j})/\tau)}
\label{eq:list-soft-p}
\end{equation}
The selector aligns its scores to the oracle distribution via
\begin{equation}
  \scalebox{0.87}{%
    $\displaystyle
    \mathcal{L}_{\text{list}}(G^{(h)})
    =\mathrm{KL}\!\left(q\,\|\,p\right)
    =\sum_{i=1}^{k} q_i\bigl(\log q_i-\log p_i\bigr)
    $}
  \label{eq:list-loss}
\end{equation}
where $\tau>0$ controls target smoothness.
Groups with $\sum_i r_i=0$ carry no loss and are skipped.
The final objective sums the listwise loss over all groups:
\begin{equation}
\mathcal{L}(\theta)
=\sum_{h}\mathcal{L}_{\text{list}}(G^{(h)})
\label{eq:total-loss}
\end{equation}
This listwise training reproduces the within-group gain ranking and concentrates probability mass on high-gain candidates, enabling top-1 expansion at inference without re-invoking the LLM. See Appendix~\ref{app:path selector} for more details.

\subsection{Text Retrieval}
\subsubsection{Company Retrieval}
To place the target company with historically comparable cases, we retrieve companies whose public descriptions are semantically similar to that of the target. The intuition is that companies sharing industry focus, product form, or market stage provide informative priors on likely financing outcomes. We use a frozen sentence encoder to embed each description and rank candidates by cosine similarity~\cite{Mikolov2013Efficient}—retaining the top-$k$ peers $\mathcal{N}_k$. To avoid temporal leakage, we only consider firms founded before the target.

\subsubsection{Investor Retrieval}
Early capital often comes with intensive screening; hence the background of the lead investor provides strong priors about a startup future trajectory. We therefore identify, for the target company, the lead investor $v^{*}$ as the one who committed the largest amount in its first disclosed financing round at time $t_{0}$.

From PitchBook entries for \(v^{*}\), we extract two types of time-stamped records and discard any entry with timestamp \(t \ge t_{0}\) to avoid leakage: empirical records 
\(\,H^{\mathrm{emp}} = \{(r_k^{\mathrm{emp}}, t_k^{\mathrm{emp}})\}\) 
and investment records 
\(\,H^{\mathrm{inv}} = \{(c_k^{\mathrm{inv}}, t_k^{\mathrm{inv}})\}\), 
where \(r_k^{\mathrm{emp}}\) denotes a role or title held by \(v^{*}\) and \(c_k^{\mathrm{inv}}\) denotes a company previously backed by \(v^{*}\). Both lists are ranked by recency and truncated to the top \(n\) items.

For every invested company $c_k^{\text{inv}}$ we attach its brief profile and historical outcome label $\mathrm{label}(c_k^{\text{inv}})\in\{\textsc{Success},\textsc{Failure}\}$. In addition, we collect static demographic attributes of $v^{*}$ (education, age, gender), denoted $A^{*}$. The resulting structured summary $\{A^{*}, H^{\text{emp}}, H^{\text{inv}}\}$ is verbalised into a investor-analysis prompt.

\subsection{Multi-agent Analysis}
We instantiate three specialist LLM agents, each mimicking a typical VC due-diligence role, to elicit complementary evidence. The \textbf{Peer-Company Analyst (PC)} agent examines the similar-company prompt built from peer-company documents. The \textbf{Investor Profile Analyst (IP)} agent reads an investor-analysis prompt that summaries the lead investor’s biography and historical portfolio. And the \textbf{Investment Chain Analyst (IC)} agent reasons over a graph-path prompt that presents information-gain–optimized chains from the investment network. Each prompt is processed by the same frozen GPT-3.5 Turbo, which outputs a binary verdict and accompanying free-form rationale. Because the LLM backbone is shared and frozen, any  differences in output reflect differences in evidence alone.

\subsection{Gating Network}
We formalize rationale fusion as a supervised weighting problem: each agent’s textual rationale is jointly embedded with the target company’s structured profile, and a lightweight gating network learns instance-specific weights to aggregate evidence for binary classification. This approach preserves richer evidence than scalar scores, yields interpretable weights linked to their supporting sentences, and maintains a frozen backbone LLM.

For each target company, we dispose of three rationales  
\(R_i\;(i \in \{\mathbf{PC},\,\mathbf{IP},\,\mathbf{IC}\})\) produced by the specialist agents.  
Each rationale is embedded using a frozen sentence encoder:  
\(r_i=f_{\text{enc}}(R_i)\in\mathbb{R}^{d_r}\).
The company’s structured attributes (industry, stage, region) are
represented by a fixed vector \(a^{*}\in\mathbb{R}^{d_a}\).
A two-layer MLP \(g_{\phi}\) scores every view conditioned on the instance.
\begin{align}
s_i &= g_{\phi}\bigl([\;r_i\;\Vert\;a^{*}]\bigr)
    \label{eq:si} \\
w_i &= \frac{\exp(s_i)}{\sum_j \exp(s_j)}
    \label{eq:wi}
\end{align}
The weights \(w_i\) vary from case to case.
The gated representation is the convex combination. 
\begin{equation}
r_{\mathrm f}
= \sum_{i} w_i\,r_i
\;\in\;\mathbb{R}^{d_r}
\label{eq:rf}
\end{equation}
which is concatenated with the attributes, is passed to another two-layer MLP \(h_{\theta}\) to obtain a task-aligned score for success prediction.
\begin{equation}
p
= \sigma\!\bigl(h_{\theta}([\;r_{\mathrm f}\Vert a^{*}])\bigr)
\;\in\;(0,1)\,
\label{eq:prediction_prob}
\end{equation}
With ground-truth label \(y\in\{0,1\}\) the gating parameters
\(\{\phi,\theta\}\) are learned by binary cross-entropy.
\begin{equation}
\mathcal{L}(\phi,\theta)
= -\,y\log p - (1-y)\log(1-p)\
\label{eq:loss}
\end{equation}
The softmax coefficients \(\{w_i\}\) therefore offer an explicit,
per-instance attribution of how much the company text, investor text, and graph path perspectives contribute to the final verdict. See Appendix~\ref{app:gating network} for more details.

\subsection{Manager Agent}
To obtain a comprehensive and human-readable final decision, the following artifacts are collated into a meta-prompt and forwarded to an additional frozen GPT-3.5 Turbo instance (the decision/manager agent): the target company's profile, the four agent predictions \(\hat{y}_{i}\), the four rationales \(R_{i}\), and the learned importance weights \(w_{i}\). Conditioned on this structured input, the decision agent produces: (i) a final binary decision  \(\hat{y}_{\text {final }} \in\{  True, False  \}\) and (ii) a natural language explanation grounded in the individual rationales and their respective weights. This architecture provides interpretable, multi-source decision-making aligned with human VC analysis practices.

\section{Experiments \& Results}
\subsection{Datasets}
\noindent\textbf{1.\ Dataset Sampling and Splitting}
We train both the Path Selector and Gate Network on subsets drawn from the original investment graph which contains a large pool of candidate companies.  To reduce computational overhead and avoid redundancy, we randomly sample 2,000 and 11,000 companies—respecting the overall success-to-failure ratio—for the Path Selector and Gate Network, respectively. Each subset is split into training, validation, and test sets in a 70 : 15 : 15 ratio, with class balance maintained across splits.

\noindent\textbf{2.\ Final evaluation}
We select 2,510 startups that successfully secured their first round of financing between October 2021 to November 2023—entirely after the LLM’s pretraining cutoff—to prevent any test instances from appearing in the pretraining corpus and avoid data leakage. This set includes 1,977 negative samples and 533 positive samples.

\subsection{Baselines}
We compare our model with state-of-the-art baselines across four categories: GNN-based methods (SHGMNN, GST), embedding-based methods (BERT Fusion), RAG-based LLM methods (RAG, GNN-RAG), and recent LLM-driven VC predictors (SSFF).

\textbf{SHGMNN}~\cite{zhang2021scalable} aggregates the heterogeneous network by meta-paths into a graph and applies a diffusion GNN with convex MAP inference in a variational EM loop to model label dependencies.
\textbf{GST}~\cite{lyu2025help} models the evolving graph of startups and investors with unsupervised graph self attention, refines embeddings via link prediction and node classification losses, and feeds monthly graph snapshots into an LSTM to predict success over five years.
\textbf{BERT Fusion}~\cite{maarouf2025fused} concatenates BERT embeddings of each startup’s Crunchbase\footnote{Crunchbase is a public platform providing comprehensive data on companies, funding rounds, investors, and market trends.} self-description with structured fundamentals and trains a lightweight neural classifier to predict success.
\textbf{Standard RAG}~\cite{lewis2020retrieval} uses a frozen dense retriever to embed queries and passages into a shared semantic space, retrieves the top $k$ similar company profiles and investor summaries, and conditions a single LLM on them via RAG Token and RAG Sequence.
\textbf{SSFF}~\cite{wang2025ssffinvestigatingllmpredictive} unites a divide-and-conquer multi-agent analyst block, an LLM-enhanced random-forest predictor with a founder-idea-fit network, and a RAG external-knowledge module to score startup prospects.
\textbf{GNN-RAG}~\cite{mavromatis2024gnn} couples a deep KGQA GNN that ranks candidate nodes and extracts shortest-path reasoning traces with an LLM that consumes those verbalized paths, yielding graph-aware RAG for KG question answering.
\begin{table*}[!t]
  \captionsetup{skip=2pt, belowskip=0pt} 
  \centering
  \caption{Performance comparison with baselines. All values are percentages (the "\%" sign is omitted). $AP@K$ indicates the monthly-averaged Precision@k.}
  \label{tab:wide_top}
  \begin{tabular}{lcccccc} 
    \toprule
    \textbf{Methods}       & \textbf{AP@5}   & \textbf{AP@10}  & \textbf{AP@20}  & \textbf{Precision} & \textbf{Recall} & \textbf{F1}   \\
    \midrule
    SHGMNN        & 25.41  & 24.56  & 26.22  & 20.65   & 82.37  & 32.97  \\
    GST           & 26.71  & 25.71  & 27.14  & 21.75   & \textbf{83.54}  & 34.51  \\
    BERT Fusion   & 24.67  & 26.67  & 25.33  & 23.63   & 24.95  & 24.27  \\
    Standard RAG  & 24.43  & 24.12  & 25.23  & 23.12   & 60.34  & 33.43  \\
    SSFF          & 28.23  & 30.02  & 28.42  & 23.23   & 69.41  & 34.81  \\
    GNN-RAG       & 29.42  & 27.53  & 27.04  & 22.81   & 71.10  & 34.54  \\
    \textbf{Ours} & \textbf{34.29}     & \textbf{32.14}   & \textbf{29.21}   & \textbf{24.32}      & 73.44     & \textbf{36.54}    \\
    \bottomrule
  \end{tabular}
  \vspace{-6pt}
\end{table*}
\subsection{Evaluation Metrics}
In evaluating binary classification tasks, standard metrics such as precision, recall, and F1 score are commonly used. However, to better reflect the practical needs of investors selecting high-potential startups, we adopt the Precision at $K$ ($P@K$) metric. $P@K$ measures the proportion of successful companies among the top $K$ model recommendations, where candidates are ranked by the model's predicted confidence. This metric is well-established in VC prediction research~\cite{sharchilev2018web,zhang2021scalable,lyu2021graph} for its ability to highlight top-performing investments. Further, to assess model performance over time, we compute the Average Precision at $K$ ($AP@K$) across monthly cohorts. A higher $AP@K$ indicates that the model consistently prioritizes successful companies, thereby offering greater practical value to investors.

\subsection{Parameter Settings}
To prevent leakage of evaluation data into the LLM’s pretraining, we use OpenAI's GPT-3.5 Turbo (knowledge cutoff: September 2021) to ensure deterministic outputs during experimentation. All training is conducted on a single NVIDIA RTX 4090 GPU with 24 GB of memory. For text embedding, we use Sentence-BERT to encode all textual inputs. Additional implementation details and parameter settings are provided in the Appendix.

\subsection{Model Performance}
All reported metrics are averaged over five independent runs of the LLM. Table~\ref{tab:wide_top} shows that our model improves AP@5 by 16.6\%, AP@10 by 16.7\%, and AP@20 by 8.0\%—measures that directly capture retrieval quality when only a handful of top candidates can be pursued in practice. Notably, our model’s AP@K gains increase as K decreases, demonstrating that higher confidence corresponds to greater accuracy in identifying promising startups—a property that is particularly valuable in real-world decision-making.
It also achieves relative gains of 5.0\% in F1 and 2.9\% in Precision over the strongest baselines, indicating a more balanced and accurate classification of success outcomes. Compared to GNN-based methods (GST and SHGMNN), which boost recall through broad structural coverage but suffer precision drops from noisy neighbors; SSFF, a recent LLM-driven VC predictor that combines multi-agent analysis and RAG yet often surfaces redundant evidence; and RAG-based approaches (Standard RAG and GNN-RAG), which ground predictions in text but ignore explicit multi-hop relational chains—MIRAGE-VC’s information-gain path retriever filters out low-value graph paths, and its multi-view gating adaptively weighs heterogeneous evidence, yielding balanced recall, higher precision, and a stronger ability to surface top-performing investment candidates. See experiments on different backbone models and train-free model in Appendix~\ref{app:backbone} and ~\ref{app:metrics}.

We then examine the results from a reasoning perspective. Analysis of 2,510 test samples shows that correctly classified startups had longer retrieved paths (4.44 vs. 3.31), suggesting richer network context aids the reasoning process (Appendix~\ref{app:err}). Appendix~\ref{app:interpret} showcases the interpretability of reasoning outputs.


\section{Ablation Study}
Table~\ref{tab:ablation} reports the ablation results. Removing the graph retrieval component significantly cuts Precision by 1.3\% and F1 by 2.5\%, highlighting the essential role of structural evidence. Both naively concatenating all 3-hop neighbors and picking paths at random degrade performance, demonstrating the superior noise-filtering capability of our path selector. Eliminating either similar company documents or the investor profiles also results in a notable performance drop, indicating their complementary value. In the fusion stage, aggregating all evidence within a single agent costs 1.4\% F1, and replacing the learnable gating mechanism with fixed weights further costs 0.6\% F1, highlighting the necessity of multi-agent fusion and adaptive gating. 
\begin{table}[htbp]
  \captionsetup{skip=1pt, belowskip=0pt}
  \centering
  \caption{Results of ablation studies.}
  \label{tab:ablation}
  \resizebox{\columnwidth}{!}{%
      \begin{tabular}{llcc}
        \toprule
        \textbf{}  & \textbf{Removed Sub-module}      & \textbf{Precision} & \textbf{F1}   \\
        \midrule
        \multirow{4}{*}{}
            & w/o Graph Retrieval         &  23.01      &  34.06        \\
            & w/o Path Selector (all)      &  22.72      &  33.29       \\
            & w/o Path Selector (random)   &  23.24      &  34.76        \\
            & w/o Similar Company         &  23.45      &  35.54       \\
            & w/o Investor Analysis       &  23.32      &  35.43       \\
        \midrule
        \multirow{2}{*}{}
            & w/o Multi-agent Fusion      &  22.97      &  35.13      \\
            & w/o Gating Network          &  24.05      &  35.94      \\
        \midrule
        \multirow{1}{*}{}
            & FULL                        &  24.32      &  36.54      \\
        \bottomrule
      \end{tabular}
  }
  \vspace{-6pt}
\end{table}

\section{Conclusion}
We present MIRAGE-VC, a multi-perspective retrieval-augmented generation framework that addresses the lack of explicit reasoning and interpretability in existing VC prediction methods. Our information-gain-driven path retriever distills massive investment networks into compact chains, enabling LLMs to perform step-by-step reasoning over investor coalitions and network positions. A learnable gating network dynamically weights different aspects of startup characteristics. Experiments show substantial improvements. Unlike prevailing graph-LLM methods that target in-graph question answering, our approach addresses off-graph prediction where graphs serve as evidence for external objectives. This paradigm extends naturally to other off-graph tasks such as recommendation and risk assessment.

\section*{Limitations}
While MIRAGE-VC pioneers the integration of graph-path retrieval with RAG in VC prediction, it exhibits the following limitations:
\begin{itemize}
    \item \textbf{Single Proprietary Dataset.} We trained and evaluated MIRAGE-VC solely on PitchBook, a proprietary dataset that cannot be publicly released. This choice reflects common practice in VC research—prior work (e.g., Venture Capital Contracts~\cite{ewens2022venture}, SHGMNN~\cite{zhang2021scalable}) similarly relies on PitchBook due to its comprehensive coverage and data quality. The primary public alternative, Crunchbase, offers an open snapshot ending in 2013\footnote{\url{https://github.com/mtwilliams/crunchbase-in-2013}}, which is too outdated for modern LLM-based prediction and would risk data leakage (many post-2013 companies likely appear in LLM pretraining corpora). Future work could enhance both generalizability and reproducibility by constructing a public benchmark: supplementing the 2013 Crunchbase snapshot with freshly crawled recent data, or collaborating with data providers to release an anonymized or time-lagged subset suitable for research.
    \item \textbf{Myopic Supervision Objective.} The current supervision scheme is local: \(\Delta\) measures only the one-hop impact of adding a single node, and our beam search optimizes these immediate gains. This greedy strategy may overlook globally optimal subgraphs or beneficial interactions among multiple nodes. To address this, we plan to explore look-ahead scoring mechanisms and sequence-level optimization techniques—such as reinforcement learning over path sequences—to capture long-term dependencies and holistic graph structures.
\end{itemize}

\section*{Ethical Considerations}
\paragraph{Data provenance and consent.}
We rely exclusively on publicly available and licensed (PitchBook) company and investor level records, using identifiers only for identity resolution. No human-subject data are collected, and we do not disclose raw documents and proprietary records.

\paragraph{Privacy, licensing, and compliance.}
All inputs come from public or licensed fields, with strict temporal ordering to prevent future-event leakage. Investor demographics (e.g., education, gender) are used solely in aggregated summaries. Users must honor the original data licenses and must not attempt re-identification.

\clearpage
\appendix
\section{Appendix}
\label{sec:appendix}
\subsection{More Related Work}
With the rise of LLMs, a variety of LLM-based financial and VC decision-support systems have been proposed~\cite{liu2023fingpt, zhang2024multimodal, ko2024can}. These systems typically rely on textual and numerical features or on simulation of real-world scenarios to improve prediction accuracy. For example, SSFF~\cite{wang2025ssffinvestigatingllmpredictive} and FinCon~\cite{yu2024fincon} systems establish a hierarchical manager–analyst collaboration mechanism, outperforming expert teams across multiple tasks; and StockGPT~\cite{mai2024stockgpt} is pre-trained on extensive quantitative stock-market data to autonomously learn price-movement patterns, yielding substantial excess returns and demonstrating the promise of generative AI in complex financial decision making. However, none of these approaches directly integrate graph-structured knowledge—such as investor–startup relationship networks—and thus they are unable to fully capture path dependencies.


\subsection{Impact of Parameters on Performance}
\begin{figure}[t]
  \centering
  \includegraphics[width=\linewidth]{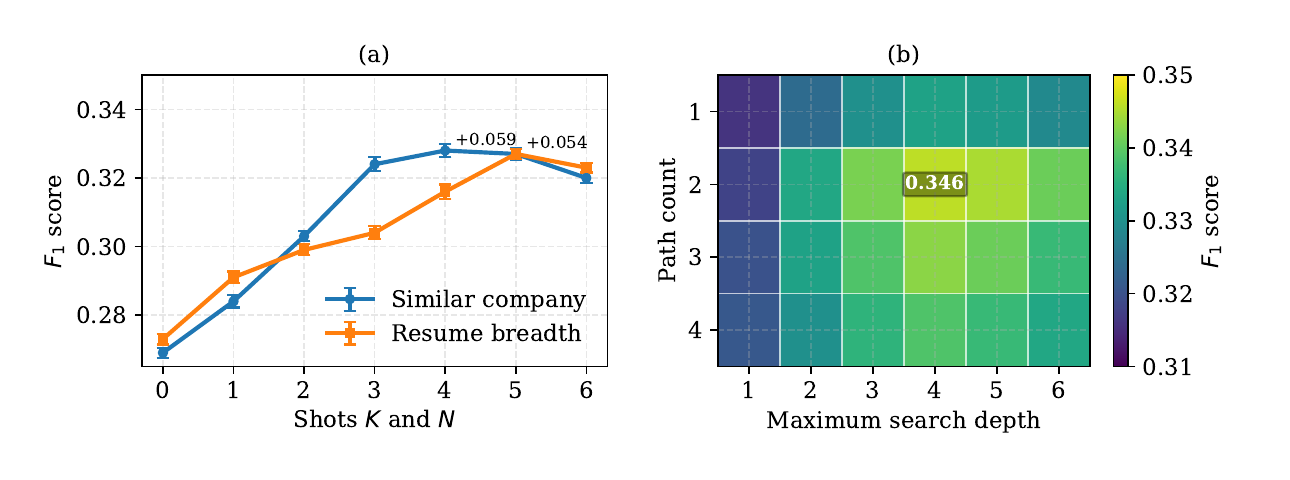}
  \caption{(a) Text retrieval: effect of the number of similar companies $K$ and resume breadth $N$ on the agents’ F$_1$. (b) Graph retrieval: effect of search depth $d_{\max}$ and path count $P$ on F$_1$.}
  \label{fig:sensitivity}
\end{figure}

We randomly sampled 1,000 companies from the non-test portion of our PitchBook data and, for each hyperparameter setting, ran each perspective’s retrieval‐and‐analysis component five times to estimate mean \(F_{1}\) scores and standard errors (Figure~\ref{fig:sensitivity}).  In the text retrieval study (Figure \ref{fig:sensitivity} (a)), we varied the number of similar‐company shots \(K\) and lead‐investor résumé entries \(N\) from 0 to 6: performance rose sharply above the zero‐shot baseline, peaked at \((K,N)=(4,5)\), and then plateaued, motivating our choice of \(K=4\), \(N=5\).  In the graph retrieval grid (Figure~\ref{fig:sensitivity} (b)), we swept maximum search depths \(d_{\max}\in\{1,\dots,6\}\) and path counts \(P\in\{1,\dots,4\}\); the highest \(F_{1}\) was observed at \((P,d_{\max})=(2,4)\), which we adopt in all main experiments.  

\subsection{Backbone Generalization and Few-Shot Comparison}\label{app:backbone}
\subsubsection{Backbone LLM Experiments}
Given the limitations of GPT-3.5-turbo, as shown in Table~\ref{tab:backbone-fewshot}, we re-ran the inference pipeline with stronger backbone LLMs, and the results show consistent improvements: With GPT-4o-mini, MIRAGE achieves a +40.1\% relative improvement in AP@5 and +11.4\% in F1. With Qwen-3 4B, MIRAGE delivers a +34.6\% improvement in AP@5 and +14.5\% in F1.
However, we note a potential data-leakage risk with newer LLMs, as their pretraining data may overlap with our test set. This is why we report results with GPT-3.5-turbo in the main paper to ensure a fair evaluation.
\subsubsection{Train-Free Few-Shot Variant}
As shown in Table~\ref{tab:backbone-fewshot}, we also implemented a train-free few-shot prompting variant as a baseline. Its performance was over 40\% lower than MIRAGE, indicating that while train-free pipelines reduce overhead, they are not yet viable when performance is a priority.
\begin{table*}[!t]
  \centering
  \caption{Backbone generalization and comparison with train-free few-shot baselines. 
All values are percentages (the ``\%'' sign is omitted). $AP@K$ denotes the monthly-averaged Precision@K.}
  \label{tab:backbone-fewshot}
  \begin{tabular}{llrrrrrr}
    \toprule
    \textbf{Backbone} & \textbf{Setting} & \textbf{AP@5} & \textbf{AP@10} & \textbf{AP@20} & \textbf{Precision} & \textbf{Recall} & \textbf{F1} \\
    \midrule
    \multirow{2}{*}{GPT-3.5-Turbo}
      & few-shot                     & 22.84 & 22.07 & 23.14 & 21.53 & 57.44 & 31.24 \\
      & + MIRAGE          & 34.29 & 32.14 & 29.21 & 24.32 & 73.44 & 36.54 \\
    \addlinespace
    \multirow{2}{*}{GPT-4o-mini}
      & few-shot                     & 25.80 & 24.73 & 24.04 & 22.81 & 66.54 & 33.53 \\
      & + MIRAGE                     & 36.35 & 33.75 & 30.38 & 25.18 & 72.21 & 37.34 \\
    \addlinespace
    \multirow{2}{*}{Qwen-3 4B (2025)}
      & few-shot                     & 21.94 & 21.14 & 21.01 & 20.04 & 63.02 & 30.41 \\
      & + MIRAGE                     & 29.53 & 28.43 & 28.14 & 23.33 & 68.55 & 34.83 \\
    \bottomrule
  \end{tabular}
\end{table*}

\subsection{Comprehensive Performance Metrics and Sensitivity Analyses}\label{app:metrics}
\subsubsection{Classification Metrics}
As detailed in Table~\ref{tab:auc_metrics}, we recalculated additional classification metrics: AUC-PR: Our method achieves a 46.9\% improvement over the random baseline and a 3.8\% improvement over the strongest baseline (SSFF). AUC-ROC: Our method achieves an 18.2\% improvement over the random baseline and a 3.0\% improvement over the strongest baseline (GNN-RAG). These results confirm that our approach consistently outperforms baselines across all key metrics.
\begin{table}[htbp]
  \centering
  \caption{Discrimination metrics across baselines. All values are in $[0,1]$.}
  \label{tab:auc_metrics}
  \begin{tabular*}{\columnwidth}{@{\extracolsep{\fill}} l  c  c }
    \toprule
    Method        & AUC-PR & AUC-ROC \\
    \midrule
    Random        & 0.241 & 0.500 \\
    SHGMNN        & 0.324 & 0.541 \\
    GST           & 0.335 & 0.556 \\
    BERT Fusion   & 0.293 & 0.543 \\
    Standard RAG  & 0.312 & 0.562 \\
    SSFF          & 0.341 & 0.571 \\
    GNN-RAG       & 0.334 & 0.574 \\
    \textbf{Ours} & \textbf{0.354} & \textbf{0.591} \\
    \bottomrule
  \end{tabular*}
\end{table}

\subsubsection{Explanation of low recall}
4.2As outlined in the paper and supported by prior VC research, real-world investment scenarios prioritize identifying the best few companies under limited resources, making AP@k the most relevant metric for evaluation. In this regard, our method consistently achieves significantly higher AP@k scores than all baselines, which reflects its effectiveness for practical application.

Regarding recall, it is important to note that the two methods with the highest recall are early GNN-based approaches. These methods achieve higher recall primarily due to:
\begin{itemize}
    \item \textbf{Over-reliance on neighborhood aggregation:} These models aggregate broad neighborhood information, which increases recall but also introduces noise.
    \item \textbf{Use of broad positive thresholds:} This further inflates recall but comes at the cost of generating many false positives, leading to much lower precision compared to other methods.
\end{itemize}
In contrast, our proposed method is designed to optimize the precision–recall trade-off and achieves the best overall performance across composite metrics. This aligns with the practical requirements of scenarios like VC prediction, where precision in selecting high-potential targets is far more critical than simply achieving high recall.

\subsubsection{Sensitivity Analysis}
As detailed in Appendix~A.2 (Figure~\ref{fig:sensitivity}), we evaluate sensitivity to the number of retrieved paths and the maximum reasoning depth. Performance is locally stable around the tuned hyperparameters: within path counts of $1$--$3$ and depths of $3$--$5$, the $F_1$ score deviates from the optimum by at most $0.003$--$0.010$. The heatmap spans $\sim 0.31$--$0.35$ and exhibits a clear ridge near $2$--$3$ paths and depth $3$--$5$, indicating low sensitivity in this region. At the extremes, shallow depth ($\leq 2$) or single-path configurations provide insufficient evidence and reduce $F_1$; conversely, overly deep settings ($\geq 6$) with many paths introduce noise and increase reasoning cost, slightly degrading performance. The results show that our method is robust to variations in hyperparameters within a broad optimal region, and achieves consistent improvements across metrics.

\subsection{Error Analysis}
\label{app:err}
\begin{figure}[htbp]
  \centering
  \includegraphics[width=\linewidth]{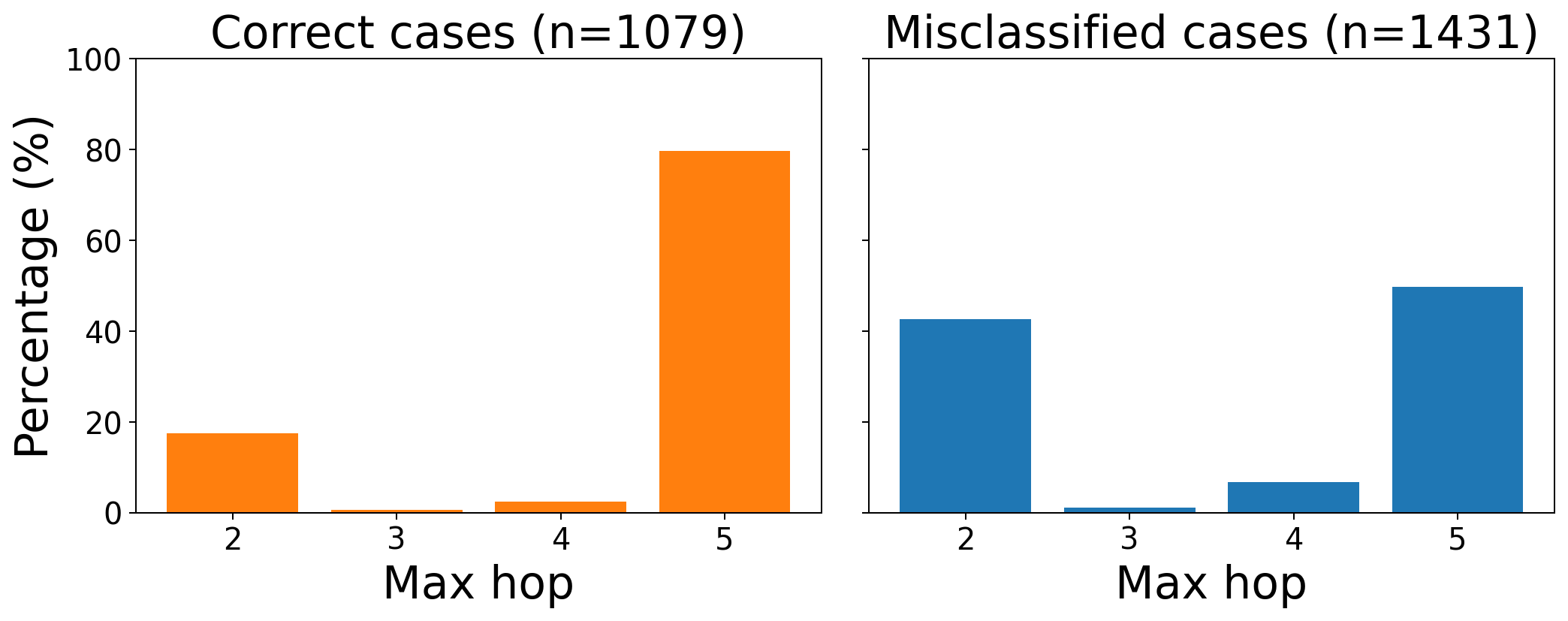}
  \caption{Distribution of path retriever maximum hop length by prediction outcome}
  \label{fig:hop-distribution}
\end{figure}

We analyze the structural characteristics that differentiate correct from incorrect predictions. Figure~\ref{fig:hop-distribution} shows that correctly predicted companies have significantly deeper accessible paths (average maximum depth 4.44 hops) compared to misclassified cases (3.31 hops). The distribution exhibits a bimodal pattern: peaks at 2 hops and 5 hops, with fewer cases at intermediate depths (3–4 hops). Two-hop cases are abundant because every company connects to at least one investor (hop 1) who has co-invested with others (hop 2), forming the minimal network structure. Five-hop cases represent our retrieval cap—networks that could extend further but are truncated at the search limit. Intermediate depths are less common because expansion depends on continuous co-investment chains: many investors' ego-networks hit dead ends after the first layer due to sparse portfolios or isolated investment patterns, causing retrieval to either stay shallow (2 hops) or jump directly to the cap (5 hops) when connections do exist. Errors concentrate in the shallow regime where limited structural evidence increases prediction ambiguity, while correct predictions benefit from the richer multi-hop context available in deeper networks.

\subsection{More Details about Data}\label{app:data}
\subsubsection{Data Sources and Time Filtering of data}
Our entity-level documents are built by concatenating descriptive tabular fields (e.g., team members, educational background, sector, stage, region) and explicitly exclude outcome/status columns tied to the prediction targets (acquisition, IPO, financing); combined with time filtering that restricts text to information available before the prediction cutoff, this design minimizes label-leakage risk. As an empirical check, we keyword-searched a random sample of 200 company profiles and 300 investor profiles for \texttt{acquisition}, \texttt{IPO}, and \texttt{financing} and found no evidence of direct leakage. We will include implementation details of the document construction and time-filtering procedures in the revised paper.
\subsubsection{Interaction Between Graph and Text Retrieval}
As mentioned above, our raw data originate from relational tables. For each entity, we build a textual profile by concatenating salient columns, followed by normalization and deduplication. This yields a corpus of entity-level documents used across modules.
\paragraph{Text retrieval.}
Text retrieval operates directly over these constructed textual profiles, selecting semantically relevant entities from the corpus (e.g., via dense or keyword retrieval).

\paragraph{Graph retrieval.}
From the relational tables we induce an investment network and retrieve key paths based on relations (e.g., company--investor--portfolio links). The nodes along the selected paths are then \emph{mapped back} to their textual profiles, which are consumed by downstream modules.

\subsection{Implementation Details}
\subsubsection{Path Selector}\label{app:path selector}
\paragraph{Binary probability of LLMs}
We follow the mainstream likelihood‑based scoring practice that normalizes the token‑level log‑likelihoods of verbalized labels (e.g., True/False) to obtain a Bernoulli probability, as adopted and analyzed in recent work on zero‑shot classification, calibration, and probability‑based prompt selection~\cite{zhou2023batch, qian2025beyond}. Given a prompt $P$, we verbalize labels as the strings ``True'' (Success) and ``False'' (Failure). For a string $w=(t_1,\dots,t_m)$, we use the string log-likelihood
\[
\log P(w\mid P)=\sum_{j=1}^{m}\log P\!\left(t_j \mid P, t_{<j}\right).
\]
Let
\[
L_T=\log P(\text{``True''}\mid P),
L_F=\log P(\text{``False''}\mid P)
\]
The success probability is obtained by two-way normalization:
\[
p=\frac{e^{L_T}}{e^{L_T}+e^{L_F}}
=\sigma\!\bigl(L_T-L_F\bigr),\quad
\sigma(x)=\tfrac{1}{1+e^{-x}}
\]
We only query log-probabilities for the target strings.
\paragraph{Training Settings}
The trade-off weight in Eq.~\eqref{eq:delta_gain},
$\lambda_{\text{conf}}\in[0,1]$, is selected on the validation split by a small grid search; Table~\ref{tab:lambda-sens} shows that performance is stable in the range $0.1\!-\!0.3$, and we therefore fix
$\lambda_{\text{conf}}=0.2$ in all experiments.
Hyperparameter is listed in Table~\ref{tab:combined-hp}.
\begin{table}[H]
  \centering
  \caption{Validation \textit{NDCG@1} (\%) under different
           confidence–trade-off weights $\lambda_{\text{conf}}$.}
  \resizebox{\columnwidth}{!}{%
    \begin{tabular}{lcccccc}
      \toprule
      $\lambda_{\text{conf}}$ & 0 & 0.10 & 0.20 & 0.30 & 0.40 & 0.50 \\
      \midrule
      NDCG@1 (\%)           & 62.7  & 63.1  & \textbf{63.4} & 63.2  & 62.9  & 62.5 \\
      \bottomrule
    \end{tabular}%
  }
  \label{tab:lambda-sens}
\end{table}
\paragraph{Evaluation and Results}
We evaluate the Path Selector with two ranking metrics. Hit@1 measures whether the selector’s top-ranked candidate matches an oracle best-gain candidate in the group. It directly reflects the success rate of our top-1 expansion policy at inference, thus aligning tightly with how the selector is actually used; a higher Hit@1 means we more often choose the maximal-gain extension. NDCG@1 normalizes the gain of the selected candidate by the maximum attainable gain in the group, yielding a score in $[0,1]$. Unlike the binary Hit@1, NDCG@1 gives partial credit when the chosen candidate is near-optimal, making it more stable under noisy or close-valued oracle gains and better for hyperparameter tuning. A random baseline is obtained by drawing an i.i.d. score from \(\mathcal{U}(0,1)\) for every candidate and applying the same evaluation procedure. Compared with random scoring, the Path Selector improves Hit@1 by \(\,+0.0879\) and NDCG@1 by \(\,+0.1851\), confirming its ability to consistently priorities expansions of higher task-specific information gain.
\begin{table}[h]
  \centering
  \caption{Selector performance on the held-out \textsc{test} split.}
  \resizebox{\columnwidth}{!}{%
    \begin{tabular}{lcc}
      \toprule
      Method  & NDCG@1 (\%) & Hit@1 (\%) \\
      \midrule
      Random $\mathcal{U}(0,1)$ & 44.92 & 33.33 \\
      \textsc{Path Selector}     & \textbf{63.43} & \textbf{42.12} \\
      \bottomrule
    \end{tabular}%
  }
  \label{tab:selector_eval}
\end{table}
\subsection{Gating Network}\label{app:gating network}
\paragraph{Data Preparation}
Each training instance consists of the three agent rationales, encoded by a frozen all-MiniLM-L6-v2 sentence encoder into \(384\)-dimensional vectors \(\{\mathbf{h}_{i}\}_{i=1}^{3}\).
Static company descriptors (industry, region, funding round) are one-hot-encoded into a \(14\)-dimensional vector \(\mathbf{c}\) and concatenated with the agent embeddings.
\paragraph{Training Settings}
Query–key projections \(\mathbf{W}_Q,\mathbf{W}_K\) are applied to each \(\mathbf{h}_{i}\) to obtain view-level attention scores. The attended view representations, together with
\(\mathbf{c}\), are fed to a two-layer MLP that outputs instance-specific weights \(\mathbf{w}\in\mathbb{R}^{3}\) (\(\sum_i w_i=1\)). The weighted sum of the three views is finally mapped by a second two-layer MLP to the task-aligned score for success prediction. All remaining hyper-parameters are summarized in Table~\ref{tab:combined-hp}.
\paragraph{Training Objective and Inference Usage}
We use an MLP only to generate an auxiliary score and train the gate under the success objective, so that the learned weights align with how much each agent (path, similar-company, lead-investor) should contribute for a given company. At inference, the gating module outputs only these weights; the final decision is made by a downstream decision agent that takes the agents’ analyses together with the learned weights.
\begin{table}[h]
    \centering
    \caption{Performance of the gating network versus a random-weight baseline
    on the held-out test set.}
    \label{tab:gate-result}
    \begin{tabular*}{\columnwidth}{@{\extracolsep{\fill}} l  c  c }
        \toprule
        Method & P (\%) & F1 (\%) \\ 
        \midrule
        Random & 19.98 & 19.49 \\
        Ours   & \textbf{23.15} & \textbf{35.13} \\
        \bottomrule
    \end{tabular*}
\end{table}
\paragraph{Evaluation and Results}
Following the main task, the gating network is assessed on Precision (P)  
and F1. As a sanity check we replace the learned weights by a uniform random choice (Random); its performance marks the chance level of selecting the most informative view.  
Table~\ref{tab:gate-result} shows that the learned gate substantially outperforms this baseline, confirming that the network has indeed captured non-trivial view–attribute interactions.
\begin{table}[h]
  \centering
  \caption{Hyper‐parameters for the Gate Network (GN) and Path Selector (PS)}
  \label{tab:combined-hp}
  \begin{tabular*}{\columnwidth}{@{\extracolsep{\fill}} l  c  c }
    \toprule
    Parameter                      & GN       & PS        \\
    \midrule
    Text vector dimension          & 384      & 384       \\
    Company key dimension    & 53       & —         \\
    Batch size                     & 256      & 256       \\
    Training epochs                & 50       & 30        \\
    Hidden width                   & 256      & 256       \\
    Optimiser                      & AdamW    & AdamW     \\
    Learning rate                  & $5\times 10^{-4}$ & $3\times 10^{-4}$ \\
    Temperature $\tau$             & —        & 0.5       \\
    \bottomrule
  \end{tabular*}
\end{table}
\subsection{Text Embedding Model Analysis}
\label{app:text_embedding}
Two pipeline components rely on a frozen sentence-encoder to obtain text
representations: (i) the \emph{graph retriever}, where the encoder embeds path
descriptions, and (ii) the \emph{gate network}, where it embeds each agent’s
generated answer.  
Here we analyse whether swapping the encoder markedly affects intermediate
metrics.  Table~\ref{tab:embed-combined} shows that across both modules the performance gap between
alternative encoders is marginal, confirming that our main results are not
sensitive to the specific choice of text-embedding model.


\begin{table}[t]
  \centering
  \caption{Impact of text encoders on the graph retriever and the gate network.}
  \label{tab:embed-combined}
  \setlength{\tabcolsep}{4pt}
  \renewcommand{\arraystretch}{1.05}
  \small
  \begin{tabularx}{\columnwidth}{
      @{}>{\raggedright\arraybackslash}p{0.25\columnwidth}
      >{\centering\arraybackslash}p{0.10\columnwidth}
      >{\centering\arraybackslash}p{0.12\columnwidth}
      >{\centering\arraybackslash}p{0.12\columnwidth}
      >{\centering\arraybackslash}p{0.12\columnwidth}
      >{\centering\arraybackslash}p{0.12\columnwidth}@{}}
    \toprule
    \multicolumn{1}{c}{\small \shortstack{Text\\encoder}} &
    \small \shortstack{Dim.} &
    \multicolumn{2}{c}{\small Graph retriever} &
    \multicolumn{2}{c}{\small Gate network} \\
    \cmidrule(lr){3-4}\cmidrule(lr){5-6}
    & & \small \shortstack{$NDCG$\\(@1,\%)} & \small \shortstack{$Hit$\\(@1,\%)}
      & \small \shortstack{$P$\\(\%)} & \small \shortstack{$F_1$\\(\%)} \\
    \midrule
    all-MiniLM-L6-v2            & 384  & 63.3 & 42.2 & 23.45 & 35.13 \\
    jina-embeddings-v2-base     & 768  & 63.4 & 42.1 & 23.12 & 35.31 \\
    e5-large-v2                 & 1024 & 63.1 & 41.9 & 23.25 & 35.25 \\
    \bottomrule
  \end{tabularx}
\end{table}

\subsection{Resource Utilization and Latency}
Our end-to-end pipeline comprises three phases: (i) retrieval via GPT-3.5 API, (ii) oracle scoring with Llama-3.1-8B and selector training, and (iii) gating network training and ablation studies. We issued approximately 40,000 GPT-3.5 requests (under 12,000 tokens each), processing \(\approx\) 480 million tokens. Locally, we generated 16,857 gain labels with Llama-3.1-8B (5,619 group expansions, all within its 8 000-token window). Training the listwise selector on an NVIDIA RTX 4090 (24 GB VRAM) for 50 epochs required about 5 minutes, and the training of gating network completed in under 10 minutes. Including hyperparameter sweeps, total GPU time was 10 GPU h. 

During inference, we noticed that LLM's usage is no higher than other well-known LLM-based forecasting systems, such as SSFF and GNN-RAG. This difference in cost is particularly insignificant when performing VC investment forecasting, a non-immediate and low-frequency forecasting task (weekly or monthly) that differs significantly from stock price forecasting. In our process, each prediction proceeds in two stages: the first stage issues three API calls, and the second stage issues a fourth call. On average, a single prediction consumes \(18,831\) input tokens and takes \(7.8\) s wall-clock. This per-instance LLM usage is operationally acceptable. In a practical setting with 1000 new target companies per month, the theoretical inference budget is \(188,310\) input tokens in total, costing about \(\$0.0192\) with GPT-4o-mini, and requiring roughly \(2\) h end-to-end runtime, which is operationally acceptable. 

\subsection{Interpretability of Multi-Agent Reasoning Outputs}
\label{app:interpret}
Using a canonical case (WhatsApp), we show how three evidence views are processed by the LLM and fused into a single binary decision (success means second round, acquisition, or IPO). The \textbf{Peer-Company}, \textbf{Investor Profile}, and \textbf{Investment Chain} analysts provide, respectively, time-consistent peer signals, lead-investor track record and alliances, and multi-hop path cues(``WhatsApp $\leftarrow$ Sequoia $\rightarrow$ Google $\leftarrow$ Kleiner Perkins $\rightarrow$ Amazon''); a \textbf{Manager Agent} combines them to produce the verdict with a concise, evidence-grounded rationale.
\subsubsection{Peer-Company Analyst}
\begin{lstlisting}[language={},basicstyle=\ttfamily\small]
### Peer-Company Analyst ###
Prediction : Success
\textbf{Analysis.} Among earlier peers in global mobile IM/OTT, the dominant exit path was acquisition after multiple rounds rather than a standalone IPO. Examples include ICQ (founded 1996; acquired by AOL in 1998), Skype (founded 2003; acquired by eBay in 2005 and later by Microsoft in 2011), and fring (founded 2006; acquired by GENBAND in 2013). Standalone IPO cases generally required a strong local monetization engine and high ARPU. In contrast, WhatsApp's product and commercial stance---minimalist, anti-advertising, and penetration-first---resembles ICQ/Skype/fring: large user scale with restrained monetization. Therefore, after its initial financing, WhatsApp is more likely to raise follow-on funding or be acquired by a platform incumbent than to pursue an independent IPO. Under the binary rule (success = second round, acquisition, or IPO), we label this case as \textbf{Success}.
\end{lstlisting}
\subsubsection{Investor Profile Analyst}
\begin{lstlisting}[language={},basicstyle=\ttfamily\small]
### Investor Profile Analyst ###
Prediction : Success
\textbf{Analysis.} WhatsApp's first external round was sole-led by Sequoia (Partner Jim Goetz). Pre-WhatsApp, Goetz had already established a successful pattern in mobile/platform assets: AdMob (mobile ads) acquired by Google for $\approx$\$750M in 2009; long-term board role at Palo Alto Networks, which IPO'd in 2012; and Clearwell Systems, acquired by Symantec in 2011. This evidences a ``concentrated, long-haul'' style and the ability to route high-penetration, low-short-term-monetization entry assets toward large follow-on rounds or strategic M\&A via superior resource orchestration. Combined with WhatsApp's product profile and user growth, this view implies a high likelihood of at least a second round and a credible path to strategic acquisition; under the binary rule, Success.
\end{lstlisting}
\subsubsection{Investment Chain Analyst}
\begin{lstlisting}[language={},basicstyle=\ttfamily\small]
### Investment Chain Analyst ###
Prediction : Success
Analysis: First, Sequoia's sustained bets and track record in early consumer-internet projects imply a mature, ``partner-style'' methodology across product refinement, growth-model validation, key-hire placement, and follow-on round syndication---creating strong signaling and improving access to subsequent financing. Second, graph linkages to established platforms such as Google/Amazon indicate structural adjacency in scaled capabilities---distribution, cloud, and advertising monetization---making it easier to obtain technical guidance, pilot opportunities, and ecosystem compliance playbooks. For a communications application that prizes reliability and low latency, such adjacency reduces engineering and compliance uncertainty during expansion and thus strengthens outside investors' confidence in sustainable growth. Third, the Sequoia $\times$ Kleiner Perkins collaboration typically confers more institutionalized capacity in follow-on introductions, acquisition-window identification, strategic-investor dialogues, and executive talent search---accelerating the opening of the next financing or M\&A window. Taken together---top-tier screening, platform adjacency, and a coordinated investor alliance---these signals reasonably raise the prior that WhatsApp will secure a next round or achieve a major liquidity event.
\end{lstlisting}
\subsubsection{Manager Agent}
\begin{lstlisting}[language={},basicstyle=\ttfamily\small]
### Manager Agent ###
Prediction : Success
Analysis: Synthesizing the three evidence sets, I materially raise the prior for a ``successful'' outcome for WhatsApp. Along the WhatsApp $\leftarrow$ Sequoia $\rightarrow$ Google $\leftarrow$ Kleiner Perkins $\rightarrow$ Amazon lineage, Sequoia's partner-style playbook plus adjacency to Google/Amazon in distribution/cloud/compliance creates reusable organizational capacity for scale-up, key hiring, and follow-on syndication---reducing engineering and regulatory uncertainty and directly improving the odds of securing subsequent rounds. The earlier-cohort baseline shows that this sector's modal exit is acquisition after multiple financings; standalone IPOs skew toward heavy monetization narratives, which contrasts with WhatsApp's minimalist, anti-ads, penetration-first posture---making strategic M\&A the more efficient medium-/long-term route. Finally, from the lead-investor vantage, Sequoia/Jim Goetz has repeatedly steered high-penetration entry assets to large follow-ons or M\&A, providing the executional lever to realize that path. The logic chain is mutually reinforcing: platform adjacency lowers uncertainty (enabling follow-ons), the sectoral precedent favors M\&A (enabling exit), and the lead-investor playbook supplies the mechanism (enabling execution). Hence the final binary decision: Success (1), with pathway ordering Follow-on financing $\gg$ Strategic acquisition $\gg$ Standalone IPO.
\end{lstlisting}

\subsection{Prompt Templates and Examples}
\label{app:prompt_templates}
This section provides the exact prompt templates used by each module and one
illustrative example per template.
\subsubsection{Company and Investor Basic Info Case}
\begin{lstlisting}[language={},basicstyle=\ttfamily\small]
### Company Profile ###
Company name    : ACME Robotics
Founded year    : 2023
Headquarters    : San Francisco, USA
Industry        : Service Robotics
Employees       : 35 (as of 2025)
Key prototype   : Compact autonomous cleaning robot for boutique hotels
Revenue status  : Pre-revenue; paid pilots scheduled Q4-2025
Funding to date : USD 3.5 M (Seed round, Jun-2024)
Lead investors  : FutureFund (Jane Doe), SeedSpark Ventures
Company overview: ACME Robotics develops AI-driven service robots that automate routine cleaning tasks in hospitality and small retail environments. The platform combines low-cost modular hardware with on-device perception and a subscription software stack, aiming to deliver pay-as-you-go automation for venues that cannot afford traditional industrial solutions.
\end{lstlisting}

\begin{lstlisting}[language={},basicstyle=\ttfamily\small]
### Lead-Investor Profile ###
Investor name: Jane Doe --- Partner @ FutureFund\\
Tenure       : 2016 -- present

Previous positions\\
$\bullet$ Senior Engineer, ABB Robotics (2008 -- 2012)---global industrial-robotics leader.\{COMPANY\_PROFILE\} (success)\\
$\bullet$ Investment Associate, TechEdge Capital (2012 -- 2016)---early-stage deep-tech VC.\{COMPANY\_PROFILE\} (success)

Focus sectors    : Robotics $\bullet$ Edge AI $\bullet$ IoT\\
Assets under mgmt: USD 1.4 B

Investment record\\
$\bullet$ RoboVac: acquired by Dyson (2021). \{COMPANY\_PROFILE\} (success)\\
$\bullet$ MechArm: IPO (2022). \{COMPANY\_PROFILE\} (success)\\
$\bullet$ NanoGrip: acquired by Bosch (2020). \{COMPANY\_PROFILE\} (success)\\
$\bullet$ ServoLink: ceased operations (2019). \{COMPANY\_PROFILE\} (failure)

Board seats : MechArm $\bullet$ FlexDroid $\bullet$ SensorX\\
Awards      : Forbes ``30 Under 40 in VC'' (2023)
\end{lstlisting}

\subsubsection{Path Analyst Prompt}
\begin{lstlisting}[language={},basicstyle=\ttfamily\small]
Role: You are a senior venture-capital analyst who excels at step-by-step
      reasoning over investment paths to judge whether a seed / angel-stage
      start-up is likely to secure Series-A funding within the next year.

You are given three blocks of information:

(1) High-value investment path retrieved for {COMPANY_NAME}:{PATH_TEXT}
(2) Company profiles appearing in the path (each with outcome labels; True = raised Series A
    within 12 months after seed/angel, False = did not):{COMPANY_PROFILES} Success/Failure:{LABELS}
(3) Investor profiles appearing in the path:{INVESTOR_PROFILES}
(4) Target company profile:{TARGET_COMPANY_PROFILE}
Task:
  • Analyse the evidence and predict whether {COMPANY_NAME} will
    raise a Series-A round within 12 months.
  • Output **exactly** in the format:

      Prediction: True/False
      Analysis: <your step-by-step reasoning>

  • If evidence is insufficient, reason cautiously but still decide.
\end{lstlisting}

\subsubsection{Company Analyst Prompt}
\begin{lstlisting}[language={},basicstyle=\ttfamily\small]
Role:
  You are a senior venture-capital analyst who excels at using information from
  industry peers (similar companies) to judge whether a seed/angel-stage target
  will secure Series-A funding within the next year.

You are given:

(1) Target company profile:{TARGET_COMPANY_PROFILE}

(2) Comparable companies (each with outcome labels; True = raised Series A
    within 12 months after seed/angel, False = did not):{COMPANY_PROFILES} Success/Failure:{LABELS}

Task:
  • Analyse the evidence and predict whether {COMPANY_NAME} will
    raise a Series-A round within 12 months.
  • Output **exactly** in the format:

      Prediction: True/False
      Analysis: <your step-by-step reasoning>

  • If evidence is insufficient, reason cautiously but still decide.
\end{lstlisting}

\subsubsection{Investor Analyst Prompt}
\begin{lstlisting}[language={},basicstyle=\ttfamily\small]
Role:
  You are a senior venture-capital analyst who specialises in evaluating a
  start-up's lead seed/angel investor record to judge whether the target can
  secure Series-A funding within the next year.

You are given:
(1) Target company profile:\{TARGET\_COMPANY\_PROFILE\}

(2) Lead-investor r\'esum\'e (prior operating roles and portfolio companies,
    each annotated as success or failure---success = the company raised Series A
    within 12 months of its seed/angel round; failure = it did not):\{INVESTOR\_PROFILE\}

Task:
  $\bullet$ Analyse how the investor's past successes and failures relate to the target
    company's sector, stage, and needs.
  $\bullet$ Predict whether the target will raise a Series-A round within 12 months.
  $\bullet$ Output \textbf{exactly} in the format:

      Prediction: True/False
      Analysis: <your step-by-step reasoning>

  $\bullet$ If evidence is insufficient, reason cautiously but still decide.
\end{lstlisting}

\subsubsection{Manager Analyst Prompt}
\begin{lstlisting}[language={},basicstyle=\ttfamily\small]
Role:
  You are a senior venture-capital analyst who excels at synthesizing other
  experts' viewpoints to decide whether a seed/angel-stage start-up will secure
  Series-A funding within the next year.

You are given:

(1) Path-analyst verdict
    $\bullet$ Prediction: \{PATH\_PREDICTION\}
    $\bullet$ Analysis  : \{PATH\_ANALYSIS\}

(2) Similar-company analyst verdict
    $\bullet$ Prediction: \{SIM\_PREDICTION\}
    $\bullet$ Analysis  : \{SIM\_ANALYSIS\}

(3) Lead-investor analyst verdict
    $\bullet$ Prediction: \{INV\_PREDICTION\}
    $\bullet$ Analysis  : \{INV\_ANALYSIS\}

(4) Aggregate-weight advice\\
    The historical importance of the three perspectives is
    \{WEIGHTS\_VECTOR\}

(5) Target company profile\\
    \{TARGET\_COMPANY\_PROFILE\}

Task:
  $\bullet$ Produce a single, final prediction on whether the target will raise a
    Series-A round within 12 months.
  $\bullet$ Output \textbf{exactly} in the format:

      Prediction: True/False
      Analysis: <your step-by-step reasoning>

  $\bullet$ If evidence is insufficient, reason cautiously but still decide.
\end{lstlisting}
\bibliography{custom}

\end{document}